\documentclass{article}
\usepackage[final]{corl_2026}
\makeatletter
\renewcommand{\@noticestring}{CoRL 2026 Workshop on Efficient Foundation Models for Real-Time Embodied AI (EEAI).}
\makeatother
\usepackage[utf8]{inputenc}
\usepackage[T1]{fontenc}
\usepackage{amsmath,amssymb,booktabs,graphicx,microtype,etoolbox,wrapfig,enumitem}
\hypersetup{hidelinks}
\pdfinfoomitdate=1 \pdftrailerid{}% reproducible PDF: no timestamps or trailer ID
\newcommand{\repourl}{https://github.com/gabhinav001/cache-or-denoise-less}
\hypersetup{pdfauthor={Abhinav Gupta, Nissan Pow},pdftitle={Cache or Denoise Less? Residual Reuse in a Flow-Matching World Action Model},pdfsubject={CoRL 2026 Workshop on Efficient Foundation Models for Real-Time Embodied AI (EEAI)},pdfkeywords={robot learning, world action models, flow matching, inference efficiency, caching}}
\newcommand{\cthree}{C$^3$ache}

\newcommand{\artifact}{\href{\repourl}{\nolinkurl{github.com/gabhinav001/cache-or-denoise-less}}}
\title{Cache or Denoise Less? Residual Reuse in a\\Flow-Matching World Action Model}
\author{
  Abhinav Gupta\\
  Independent Researcher\\
  \texttt{gabhinav001@gmail.com}
  \And
  Nissan Pow\\
  Independent Researcher\\
  \texttt{nissan.pow@gmail.com}
}
\begin{document}
\maketitle

\begin{abstract}
Flow-matching robot policies generate each action chunk through repeated network evaluations, one per denoising step.
Cross-cycle caches such as \cthree{} reuse computation from earlier replanning calls, but whether they beat simply taking fewer steps is unclear.
We compare the two in Fast-WAM, a 6B-parameter world action model, on LIBERO-Goal.
Two uncached steps keep the success of ten: on all 500 initial states they are non-inferior within two percentage points, for one checkpoint and noise seed.
Under PyTorch's default compile mode, they also cut a replanning call on an RTX A6000 from 202 to 92\,ms.
This leaves little room for caching. A cache that keeps two fresh steps cannot be faster than two uncached steps: every call still encodes the images, runs the video expert and computes those two steps. Caching the first of two steps, refreshed every fourth call, saves at most 11--14\% of a call by our cost model.
Cached policies are also fragile. Their losses tracked the share of the flow path the cached steps cover: under Fast-WAM's current time shift, \cthree{}'s eight cached steps cover 80\% of the path instead of 44\%, and the cached policy succeeds on 63\% of states against 95\% uncached. Resampling the noise on every call cost caches covering 60\% or more of the path 14--40 points.
These results argue for fewer-step baselines and for testing reuse under the deployed sampler and noise settings.
We release our evaluation harness, logs and a checklist.
\end{abstract}

\keywords{Robot policies, world action models, flow matching, inference efficiency, caching}

\section{Introduction}
Diffusion and flow-matching policies~\citep{diffusionpolicy2023,pi0,pi05} and world action models~\citep{fastwam2026,dreamzero2026} produce each action chunk by integrating a learned flow from noise, one network evaluation per step.
A ten-step replanning call of Fast-WAM~\citep{fastwam2026} takes about 200\,ms on a desktop GPU, four control steps of a 20\,Hz robot, which must pause for each new plan or keep executing an old one.
Distillation~\citep{consistencypolicy2024,onestepdp2024,flashwam2026} and coarser schedules~\citep{sbintuitions2026} cut the number of steps. Caches reuse computation within a denoising trajectory~\citep{deepcache2024,wimbauer2024cache,teacache2025,magcache2025,taylorseer2025,bac2026,efficientvla2025}, across chunks~\citep{falcon2025,streamingdp2024,xcache2026} or across control cycles~\citep{vlacache2025,c3ache2026}.

\cthree{}~\citep{c3ache2026} is a cross-cycle cache for Fast-WAM. A refresh call runs all ten steps and stores the residual that the action transformer adds at each of the first seven or eight. Until the next refresh, every call adds the stored residuals instead of recomputing those steps and runs only the rest in full.
It reports up to 2.5$\times$ faster inference than ten steps at unchanged success. That baseline has since moved: Yang~\citep{sbintuitions2026} found that two undistilled steps already reach 95.8\% success on LIBERO-Goal.

A cache now has to beat two uncached steps. We compare caches with that baseline on Fast-WAM's two released LIBERO checkpoints and describe two conditions under which caches lose success. X-Cache~\citep{xcache2026} reuses residuals across the chunks of a few-step driving world model, gating each block on its input and always recomputing the chunks that write its key--value cache. It notes that resampled noise leaves consecutive chunks dissimilar at the first denoising step, and compares with its own uncached model.
Two uncached steps are non-inferior to ten, and every cache that keeps two full steps is slower, because a reuse call still encodes the images, runs the video expert and computes the last two steps (Sec.~\ref{sec:frontier}).
A cache's losses track the share of the flow path its cached steps cover, a share that Fast-WAM's latest time shift raised from 44\% to 80\% for eight cached steps, and the cached policy falls to 63\% success (Sec.~\ref{sec:shift}).
The losses grow when the noise is resampled on every call, and a pre-registered control points to the latent changing between calls.
Sec.~\ref{sec:practical} collects the lessons in a checklist, and code and logs are at \artifact.

\section{Setup}
\label{sec:setup}

\begin{figure}[t]
\centering
\includegraphics[width=0.70\linewidth]{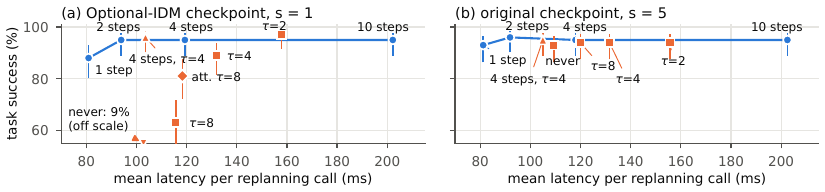}
\vspace{-2pt}
\caption{Success against mean latency per call on 100 LIBERO-Goal states (one RTX A6000). Blue: uncached, 1--10 steps. Orange: caches refreshed every $\tau$ calls (squares: 8 of 10 steps cached; diamond: attenuated; triangle: 2 of 4 cached). Bars: 95\% Wilson intervals.}
\label{fig:frontier}
\end{figure}

\paragraph{Policy.}
Fast-WAM~\citep{fastwam2026} pairs a 5B video transformer with a 1B action transformer.
Each replanning call encodes two camera images, runs the video expert once to build a key--value cache, and denoises a chunk of 32 actions with the 30-block action expert, which attends to that cache. The robot executes 10 of them (0.5\,s at 20\,Hz) and replans.
Denoising integrates a flow from noise ($\sigma{=}1$) to actions ($\sigma{=}0$)~\citep{flowmatching2023} with $K$ Euler steps on the grid
{\setlength{\abovedisplayskip}{3pt}\setlength{\belowdisplayskip}{3pt}
\begin{equation}
\sigma_k = \frac{s\,u_k}{1+(s-1)\,u_k}, \qquad u_k = 1-k/K, \qquad k = 0,\dots,K,
\label{eq:shift}
\end{equation}}%
where a larger shift $s$~\citep{sd3} keeps more steps at high noise.
When the first $m$ steps are cached, we call $1-\sigma_m$ the cache's \emph{share} of the flow path.
Both released LIBERO checkpoints use $K{=}10$; the \emph{original} runs at $s{=}5$, the shift its release specifies.
The \emph{Optional-IDM} checkpoint can infer actions from imagined frames; we run it in first-frame mode at $s{=}1$, the current shift (commit \texttt{0050a44}).

\paragraph{Cache.}
We implement \cthree{}'s residual cache~\citep{c3ache2026} over the whole 30-block stack.
A \emph{refresh} call runs normally and stores the stack's residual $R_k = h_{30} - h_0$ at each cached step $k$, and a later \emph{reuse} call skips the stack there and sets $h_{30} = h_0 + R_k$.
We cache steps \mbox{0--7} of 10, \cthree{}'s longer range, and refresh them every $\tau$ calls, by default $\tau{=}8$. With $\tau{=}0$ the cache is filled only when an episode starts, \cthree{}'s headline setting (2.51$\times$ on an A100).
An \emph{attenuated} variant scales each block's stale residual by a least-squares weight fitted on replayed demonstrations.
Before each collection, a gate reuses the cache on the call that filled it and checks the actions against uncached ones (Appendix~\ref{app:gate}).

\paragraph{Protocol.}
All evaluation is closed-loop in LIBERO simulation~\citep{libero2023}, on 10 initial states for each of the 10 LIBERO-Goal tasks (100 states), or on the first 5 per task (50 states) for the shift and noise experiments and LIBERO-Long.
Each configuration runs in its own Python process on an RTX A6000, and Table~\ref{tab:frontier} comes from a single machine.
Fast-WAM reseeds its noise generator identically on every call, so rollouts nearly repeat across machines (7 of 11,725 paired outcomes changed over thirteen machines), and we compare success across machines but latency only within one.
Latency is the CUDA-synchronized time of a replanning call under PyTorch's default compile mode, excluding each task's first episode, which compiles; we also time uncached schedules with CUDA graphs.
Success is compared state by state with exact McNemar tests, bootstrap intervals, and Tango score intervals for non-inferiority within five points~\citep{tango1998}, a margin we fixed beforehand as roughly what 100 paired states resolve near 95\% success. Since it lets a 5\% failure rate double, we report each interval.
Holm's correction~\citep{holm1979} covers McNemar tests of equal success for the eleven primary contrasts of Sec.~\ref{sec:frontier}, listed in Appendix~\ref{app:contrasts}; other tests are uncorrected and exploratory.

\section{Fewer steps or a cache?}
\label{sec:frontier}
\begin{table}[t]
\centering
\renewcommand{\arraystretch}{0.8}
\caption{Success and latency per call on 100 LIBERO-Goal states (one RTX A6000), Optional-IDM (IDM) at $s{=}1$ and the original at $s{=}5$, under PyTorch's default compile mode. Cache latencies include each episode's opening refresh. Means average episodes' mean call times; p95 pools all calls.}
\label{tab:frontier}
\footnotesize
\begin{tabular}{@{}lrrrrrrrrr@{}}
\toprule
 & & \multicolumn{2}{c}{path cached (\%)} & \multicolumn{2}{c}{success (\%)} & \multicolumn{2}{c}{mean latency (ms)} & \multicolumn{2}{c}{p95 latency (ms)} \\
\cmidrule(lr){3-4}\cmidrule(lr){5-6}\cmidrule(lr){7-8}\cmidrule(l){9-10}
Configuration & $K$ & IDM & orig. & IDM & orig. & IDM & orig. & IDM & orig. \\
\midrule
Uncached & 1 & -- & -- & 88 & 93 & 81 & 81 & 82 & 82 \\
Uncached & 2 & -- & -- & 95 & 96 & 94 & 92 & 96 & 95 \\
Uncached & 4 & -- & -- & 95 & 95 & 119 & 118 & 125 & 122 \\
Uncached & 10 & -- & -- & 95 & 95 & 202 & 202 & 209 & 209 \\
\midrule
Cache, $\tau{=}4$ & 4 & 50 & 17 & 96 & 95 & 104 & 105 & 123 & 124 \\
Cache, $\tau{=}2$ & 10 & 80 & 44 & 97 & 94 & 158 & 156 & 212 & 208 \\
Cache, $\tau{=}4$ & 10 & 80 & 44 & 89 & 94 & 132 & 132 & 214 & 212 \\
Cache, $\tau{=}8$ & 10 & 80 & 44 & 63 & 94 & 116 & 120 & 209 & 214 \\
Cache, never refreshed & 10 & 80 & 44 & 9 & 93 & 103 & 109 & 106 & 208 \\
Attenuated cache, $\tau{=}8$ & 10 & 80 & 44 & 81 & -- & 118 & -- & 221 & -- \\
\bottomrule
\end{tabular}

\end{table}

Fast-WAM does not need ten denoising steps on LIBERO-Goal (Fig.~\ref{fig:frontier}, Table~\ref{tab:frontier}). Two uncached steps succeed on 95 and 96 of 100 states (Optional-IDM and original), against 95 and 95 with ten, at 94 and 92\,ms per call instead of 202, and both are non-inferior within five points (Tango 90\% intervals $[-4.1, 4.1]$ and $[-2.6, 5.0]$).
On all 500 initial states, in an earlier Optional-IDM collection that our 100 states reproduce exactly, two steps succeed on 96.6\% and ten on 96.4\% (90\% interval $[-1.4, 1.8]$), non-inferior even at a two-point margin. That is the firmer result, since Optional-IDM's 100-state interval falls to $-6.5$ without its most favorable task (Appendix~\ref{app:pertask}). Two steps also clear a two-point margin on 250 states under 1.5\,cm camera jitter (70.0 against 64.8\%, $[1.4, 9.3]$). In a synchronous loop the faster call shortens each chunk only 1.18-fold (594 against 702\,ms).

A cache that keeps two fresh steps cannot be faster than two uncached steps. Its reuse call still encodes the images, runs the video expert and takes the two fresh steps, so it costs a two-step call plus bookkeeping (100 vs.\ 94\,ms), and a refresh call costs the full schedule (210\,ms).
Fitting a fixed cost $F$ plus a per-step cost $a$ to the uncached calls gives $F \approx 4.8a$--$4.9a$, so about 70\% of a two-step call is work no action cache reduces.
Upstream CUDA graphs cut two- and ten-step calls to 86 and 143\,ms, while the fastest cache for the original checkpoint, which our implementation cannot run under graphs, takes 105\,ms. A graph-compatible cache would still pay for a two-step call on every reuse.
No cache is significantly more successful either, though the closest, Optional-IDM's $\tau{=}2$ cache (97 against 95 states, at 158 instead of 94\,ms), could still be a few points better ($[-6.9, 2.5]$).
With the original checkpoint the $\tau{=}8$ cache costs one point (\cthree{} reports none), while Optional-IDM degrades as refreshes grow rarer. After Holm's correction, two uncached steps beat its $\tau{=}8$ cache by 32 points ($p{<}10^{-4}$) and its attenuated version by 14 ($p{=}0.02$).

Once a schedule has two steps, most of its cost lies outside the action expert. Serving the first of the two steps from a cache refreshed every fourth call ($\tau{=}4$) skips one action-expert pass on three calls in four. The cost model, assuming a cached step costs nothing and the rest of the call is unchanged, bounds the saving at 11\% of a call on the Table~\ref{tab:frontier} machine (14\% with the follow-up machine's fit, and 6\% if both calls ran under CUDA graphs). Rarer refreshes raise the ceiling only toward 15--19\%.
This composed cache succeeded on 94 and 95 of 100 states against 95 and 96 for two uncached steps (Tango 90\% interval $[-5.0, 2.6]$ for both). It saved 10\% of a call on an idle A6000 machine (95\% interval $[-1.8, 22.2]\%$ over three rounds) and 2--9\% on loaded ones, so neither difference is resolved.

\textbf{Delayed execution.} We also emulated asynchronous execution: while a call runs, the robot executes the previous chunk, so a call taking $d$ control steps replaces the first $d$ actions of its chunk with older plans. The delay $d$ is fixed per configuration, the mean call time of Table~\ref{tab:frontier} rounded up to 50\,ms steps. Two steps at delay 2 succeeded on 48 of 50 states and ten steps at delay 5 on 43, the direction we predicted, but the exact test does not resolve the gap ($p{=}0.12$). Against no delay, delay 5 cost two steps 12 points and ten steps 10 ($p{=}0.11$ and 0.18), and delays of 2--4, including the 3 that CUDA graphs give ten steps, changed ten-step success by at most 6 (Appendix~\ref{app:graphs}).

\section{When does a cache lose success?}
\label{sec:shift}
\begin{wrapfigure}{r}{0.46\linewidth}
\vspace{-10pt}
\centering
\includegraphics[width=\linewidth]{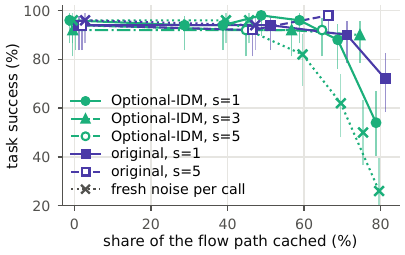}
\vspace{-6pt}
\caption{\footnotesize Success against the cached share of the flow path ($\tau{=}8$, 50 states; 0\% is uncached). Triangles: $s{=}3$. Crosses: fresh noise per call. Bars: 95\% Wilson intervals.}
\label{fig:shift}
\vspace{-6pt}
\end{wrapfigure}

The never-refreshed cache of Table~\ref{tab:frontier} succeeds on 9\% of states with Optional-IDM at $s{=}1$ but on 93\% with the original at $s{=}5$. To separate checkpoint from shift, we ran each checkpoint at both shifts with the default $\tau{=}8$ cache on the first 50 states (Fig.~\ref{fig:shift}).
Moving Optional-IDM from $s{=}1$ to $s{=}5$ raises its cached success from 54 to 92\%, and moving the original from $s{=}5$ to $s{=}1$ lowers it from 92 to 72\%. Both changes exceed the shift's effect on uncached success (at most two points), by 40 and 20 points (95\% intervals $[24, 56]$ and $[6, 34]$).
The checkpoint matters too: cached at $s{=}1$, the original keeps 18 points more than Optional-IDM (72 vs.\ 54\%, $p{=}0.02$), against two points uncached.

At $s{=}5$ the grid of Eq.~\ref{eq:shift} crowds its early points near $\sigma{=}1$, so caching steps \mbox{0--7} integrates only 44\% of the flow path, while at $s{=}1$ the same steps integrate 80\%.
Across cached-step counts and a third shift, $s{=}3$, no fixed-noise cache covering 64\% of the path or less lost more than two points.
At 70--75\% the losses were mixed (2--8 points, intervals reaching $-18.4$), and at 80\% both checkpoints lost 22--42.
The eight cached steps that fail at $s{=}1$ cover 57\% at $s{=}3$ and succeed on 92\% of states (19 to 0 discordant pairs), so a step-counting rule changes meaning with the sampler. The share describes these runs without isolating a cause, since a new shift also moves the fresh steps that could correct a cached error (Appendix~\ref{app:share}).
For an unseen shift, $s{=}2$, the observed losses (2 and 26 points at 54 and 82\% of the path) agreed with predictions committed before the runs, but only as point estimates: 50 states establish neither bound (Appendix~\ref{app:heldout}), so we claim no safe boundary.

\textbf{The noise latent.} Fast-WAM starts every call from the same latent. In exploratory runs that resample it on every call, uncached policies stayed within two points but caches covering 60\% or more of the path lost 14--40 (uncorrected $p \le 0.039$), and a pre-registered second stream repeated the 14-point loss at 60\% (pooled $p{=}0.006$ over initial states).
Fresh noise changes the latent both across episodes and between calls; a pre-registered control keeps each episode's first fresh latent for all of its calls. The 60 and 80\% caches then succeeded on 96 and 64\% of states, against 96 and 54\% with the run's fixed latent and 82 and 26\% with fresh noise ($p{=}0.065$ and $p{<}0.001$ against fresh noise). For Optional-IDM at these two shares, then, the loss appears to come mainly from reusing residuals across calls that start from different latents.
When uncached rollouts are replayed under each cache (Appendix~\ref{app:screen}), stored residuals drift by 56--66\% between calls with a fixed latent, 65\% with a per-episode latent and 85--97\% with fresh noise. Fits on the resulting action error and on the share predicted 61 and 62\% for a held-out configuration that reached 20\%.

\section{A checklist for evaluating inference accelerators}
\label{sec:practical}
\begin{enumerate}[leftmargin=*,nosep,label=\arabic*.]
\item \textbf{Compare with fewer steps.} A cache with $k$ fresh steps ran slower than $k$ uncached steps.
\item \textbf{State reuse rules in noise level, and re-test after sampler changes.} Eight cached steps covered 44, 57 and 80\% of the path at $s{=}5$, 3 and 1.
\item \textbf{Test with deployed noise.} Fixed noise hid losses of 14--40 points.
\item \textbf{Isolate configurations; time them on one machine.} Sharing a process changed 17--19 of 20 pilot trajectories, and latency varied 2.4-fold across machines.
\item \textbf{Measure success under delay.} A synchronous simulator cannot show what latency costs.
\end{enumerate}

\smallskip\noindent\textbf{Limitations.}
All runs are in simulation, with fixed delays in whole control steps and no latency jitter, concurrency or chunk blending~\citep{rtc2025}.
We studied one architecture, two checkpoints, mostly ten LIBERO-Goal tasks and one noise seed (one latent at 500 states), desktop GPUs and whole-step caches without CUDA graphs, and measured neither memory nor energy.

\clearpage
{\small
\setlength{\bibsep}{0.2pt plus 0.2pt minus 0.2pt}
\bibliography{refs}
}

\clearpage
\appendix
\renewcommand{\thetable}{A\arabic{table}}
\setcounter{table}{0}
\renewcommand{\thefigure}{A\arabic{figure}}
\setcounter{figure}{0}
\section{Additional results}
\label{app:supp}
\noindent Every value in the tables of this appendix is generated from \nolinkurl{results/budget_shift/summary.json}, the analysis snapshot in our repository.

\subsection{Primary contrasts}
\label{app:contrasts}
Table~\ref{tab:app-contrasts} lists the eleven contrasts that Holm's correction covers in Sec.~\ref{sec:frontier}, on the 100 LIBERO-Goal states of Table~\ref{tab:frontier}.
Only the three contrasts involving the $\tau{=}8$ cache and its attenuated version with Optional-IDM survive the correction.
These are tests of equal success. Non-inferiority of two against ten steps is a separate one-sided test; at a three-point margin it still holds for the original checkpoint ($[-2.6, 5.0]$) but not for Optional-IDM on 100 states ($[-4.1, 4.1]$), whose 500-state interval, $[-1.4, 1.8]$, clears even two points.

\begin{table}[!ht]
\centering
\caption{The eleven primary contrasts of Sec.~\ref{sec:frontier} (100 LIBERO-Goal states). A and B are success rates, the difference A$-$B is paired by state and given in points, $p$ is an exact McNemar test, and Holm $p$ is adjusted over the family.}
\label{tab:app-contrasts}
\footnotesize
\begin{tabular}{@{}llrrrrrr@{}}
\toprule
Ckpt. & Contrast (A vs.\ B) & A (\%) & B (\%) & A$-$B & Tango 90\% & $p$ & Holm $p$ \\
\midrule
IDM & 2 vs.\ 10 steps & 95 & 95 & 0 & $[-4.1, 4.1]$ & 1.00 & 1.00 \\
IDM & 2 steps vs.\ cache $\tau{=}8$ & 95 & 63 & +32 & $[23.3, 40.8]$ & $<10^{-4}$ & $<10^{-4}$ \\
IDM & 2 steps vs.\ attenuated $\tau{=}8$ & 95 & 81 & +14 & $[7.2, 21.6]$ & 0.0026 & 0.023 \\
IDM & 2 steps vs.\ cache $\tau{=}2$ & 95 & 97 & -2 & $[-6.9, 2.5]$ & 0.69 & 1.00 \\
IDM & 2 steps vs.\ 4-step cache $\tau{=}4$ & 95 & 96 & -1 & $[-4.4, 1.7]$ & 1.00 & 1.00 \\
IDM & attenuated vs.\ plain $\tau{=}8$ & 81 & 63 & +18 & $[9.1, 27.0]$ & 0.0021 & 0.021 \\
\midrule
orig. & 2 vs.\ 10 steps & 96 & 95 & +1 & $[-2.6, 5.0]$ & 1.00 & 1.00 \\
orig. & 2 steps vs.\ cache $\tau{=}8$ & 96 & 94 & +2 & $[-1.7, 6.4]$ & 0.62 & 1.00 \\
orig. & 2 steps vs.\ never refreshed & 96 & 93 & +3 & $[-0.9, 7.8]$ & 0.38 & 1.00 \\
orig. & 2 steps vs.\ 4-step cache $\tau{=}4$ & 96 & 95 & +1 & $[-3.3, 5.6]$ & 1.00 & 1.00 \\
orig. & 1 vs.\ 10 steps & 93 & 95 & -2 & $[-6.4, 1.7]$ & 0.62 & 1.00 \\
\bottomrule
\end{tabular}

\end{table}

\subsection{Per-task results}
\label{app:pertask}
Table~\ref{tab:app-per-task} gives the successes behind Table~\ref{tab:frontier} task by task. On every task, two and ten steps differ by at most two of ten states for either checkpoint, so the overall agreement does not hide large opposite swings. Single tasks do matter for the interval: leaving out one task at a time, the lowest Tango 90\% bound of two against ten steps is $-3.8$ for the original checkpoint but $-6.5$ for Optional-IDM, reached when the task where two steps gained two states is the one left out. For Optional-IDM the 500-state comparison is the firmer evidence. The $\tau{=}8$ cache's losses with Optional-IDM come from five tasks, which it completes on 1--5 of ten states, while it loses at most two states on each of the other five.

\subsection{Cached share and fresh noise}
\label{app:share}
The share is the part of the noise interval that the cached steps span, not a length along the learned trajectory.
Table~\ref{tab:app-share} gives every point of Fig.~\ref{fig:shift}. The $s{=}3$ points and the 70--75\% shares come from a follow-up machine, which also reproduced the outcomes of 330 episodes that earlier machines had run.
Under fixed noise, no cache covering 64\% of the path or less lost more than two points, the caches at 70--75\% lost 2--8, and those at 80\% lost 22--42.
Under fresh noise, losses of 14 points or more appeared from a share of 60\% on. At similar shares on different grids, success agreed below two-thirds of the path (92--96\% at 54--64\% for $s{=}1$, 2, 3 and 5) but not above it: caching 82\% of the path at $s{=}2$ kept 70\%, against 54\% for 80\% at $s{=}1$ and 90\% for 75\% at $s{=}3$.
Two fresh steps at $s{=}5$ integrate 56\% of the path, consistent with the original checkpoint's tolerance of even a cache filled once per episode (93\% in Table~\ref{tab:frontier}).

\subsection{Held-out shift and a second noise stream}
\label{app:heldout}
The predictions of Sec.~\ref{sec:shift} are stated in \texttt{repro/experiments/heldout.sh}, which we committed before renting the machine: at most four points lost with the first seven of ten steps cached at $s{=}2$ (54\% of the path) and at least 20 with the first nine (82\%). At $s{=}2$, ten uncached steps succeeded on 48 of 50 states, the cache over the first seven steps on 47 (Tango 90\% interval $[-9.7, 5.1]$), and the cache over the first nine on 35, losing 15 states and gaining 2 ($[-38.5, -13.7]$). With run seed 7, the 60\% cache succeeded on 45 states under fixed noise and 38 under fresh noise ($[-25.4, -3.5]$; $p{=}0.065$ on its own), against 48 and 41 with seed 42. Pooled over both seeds, the fresh-noise loss has $p{=}0.003$ with the 100 state-seed pairs as the unit and $p{=}0.006$ with the 50 initial states as the unit, from an exact sign-flip test on each state's summed difference. Fresh noise also gives every episode its own first latent, whereas fixed noise reuses one latent for the whole run. A second fixed latent (seed 7) kept the 60\% cache at 90\%, near the 96\% of seed 42, so the advantage of fixed noise is not tied to one draw. The per-episode latent control below separates reuse across calls from the choice of latent.

\begin{table}[!ht]
\centering
\caption{Every point of Fig.~\ref{fig:shift} ($\tau{=}8$, 50 LIBERO-Goal states). \emph{Cached}: leading steps of ten served from the cache. $\sigma$ \emph{fresh}: noise level at which fresh computation resumes. \emph{Fixed} and \emph{fresh}: success (\%) under Fast-WAM's fixed initial noise and with noise resampled on every call. $p$: exact McNemar test of fresh against fixed noise, uncorrected.}
\label{tab:app-share}
\footnotesize
\begin{tabular}{@{}lrrrrrrr@{}}
\toprule
Ckpt. & $s$ & cached & share (\%) & $\sigma$ fresh & fixed (\%) & fresh (\%) & $p$ \\
\midrule
IDM & 1 & 0 & 0 & -- & 96 & 96 & 1.00 \\
IDM & 1 & 3 & 30 & 0.70 & 94 & -- & -- \\
IDM & 1 & 4 & 40 & 0.60 & 94 & 96 & 1.00 \\
IDM & 1 & 5 & 50 & 0.50 & 98 & -- & -- \\
IDM & 1 & 6 & 60 & 0.40 & 96 & 82 & 0.039 \\
IDM & 1 & 7 & 70 & 0.30 & 88 & 62 & 0.0023 \\
IDM & 1 & 8 & 80 & 0.20 & 54 & 26 & 0.0013 \\
IDM & 3 & 0 & 0 & -- & 92 & -- & -- \\
IDM & 3 & 8 & 57 & 0.43 & 92 & -- & -- \\
IDM & 3 & 9 & 75 & 0.25 & 90 & 50 & $<10^{-4}$ \\
IDM & 5 & 0 & 0 & -- & 94 & -- & -- \\
IDM & 5 & 8 & 44 & 0.56 & 92 & 96 & 0.62 \\
IDM & 5 & 9 & 64 & 0.36 & 92 & -- & -- \\
orig. & 1 & 0 & 0 & -- & 94 & -- & -- \\
orig. & 1 & 5 & 50 & 0.50 & 94 & -- & -- \\
orig. & 1 & 7 & 70 & 0.30 & 90 & -- & -- \\
orig. & 1 & 8 & 80 & 0.20 & 72 & -- & -- \\
orig. & 5 & 0 & 0 & -- & 94 & 96 & 1.00 \\
orig. & 5 & 8 & 44 & 0.56 & 92 & 94 & 1.00 \\
orig. & 5 & 9 & 64 & 0.36 & 98 & -- & -- \\
\bottomrule
\end{tabular}

\end{table}

\subsection{Per-episode latent and delayed execution}
\label{app:latentdelay}
The predictions P9--P11 are stated in \texttt{repro/experiments/latent\_delay.sh}, which we committed with the analysis that tests them before renting the machine. All runs use the 50 states of Fig.~\ref{fig:shift} and Optional-IDM at $s{=}1$ unless noted.
In the latent control, Fast-WAM is seeded once per episode with the seed that the fresh-noise arms reseed their generator with. Fast-WAM draws the action latent first in every call, so every call of an episode starts from the latent of that episode's first fresh-noise call, and each run records a digest of every episode's latent.
P9 predicted that this per-episode latent would keep the 60 and 80\% caches at least eight points above fresh noise, and it did (+14 and +38). P10 predicted success within ten points of the run's fixed latent, which held, at 80\% only at its boundary (0 and +10).
Screening the uncached policy under the per-episode latent, all calls of each episode started from one latent, and the stored residuals differed from the current call's by 65\%, against 63 and 85\% for the same policy under a fixed latent and under fresh noise, with $D$ at eight cached steps 0.40 against 0.40 and 0.52.
Under delayed execution, the robot executes the previous chunk's actions for the first $d$ control steps after a call and the new chunk's actions $d$ to 9 after that, and holds still before an episode's first chunk arrives. The delays, 2 and 5 steps, are the two- and ten-step call times of Table~\ref{tab:frontier} rounded up to 50\,ms steps.
P11 predicted that two steps at delay 2 would succeed at least as often as ten at delay 5, and they did (48 against 43 of 50 states). As stated, P9--P11 bound point estimates. By exact McNemar tests at the 5\% level, these states resolve the per-episode latent's gain over fresh noise at 80\% of the path and the cost of delay 5 against delay 2 for two steps, but not the gain at 60\% or the P11 comparison, whose 90\% Tango intervals exclude zero while their exact tests do not (Table~\ref{tab:app-latent-delay}). Table~\ref{tab:app-latent-delay} gives every contrast, including the original checkpoint's $\tau{=}8$ cache, delayed 5 steps after refresh calls and 2 after reuse calls, which succeeded on 88\% of states against 94\% for its two uncached steps at delay 2 and 92\% without delay.

\subsection{Delays of CUDA-graph execution}
\label{app:graphs}
Under upstream CUDA graphs, two- and ten-step calls take 86 and 143\,ms on the Table~\ref{tab:frontier} machine, delays of 2 and 3 control steps. A follow-up committed with its analysis before the runs (\texttt{repro/experiments/delay\_graphs.sh}), without a prediction, ran ten steps at delay 3 for both checkpoints and at delay 4 for Optional-IDM. Ten steps with Optional-IDM then succeeded on 96, 96, 90, 94 and 86\% of the 50 states at delays of 0, 2, 3, 4 and 5, and with the original checkpoint on 94\% at delays 0 and 3. Against two steps at delay 2, ten steps at delay 3 lost 6 points with Optional-IDM ($p{=}0.25$) and none with the original checkpoint (Table~\ref{tab:app-latent-delay}). Only delay 5 cost both schedules consistently. In a synchronous loop, two steps finish a chunk 1.18 times faster than ten (594 against 702\,ms at 20\,Hz).

\subsection{Other regimes}
\label{app:regimes}
Table~\ref{tab:app-regimes} compares two and ten uncached steps beyond the 100 states of Table~\ref{tab:frontier}.
A pilot of ten episodes per magnitude chose the camera jitter: ten steps succeeded on 9, 9 and 5 of 10 episodes at 7.5, 10 and 15\,mm.
Most runs share one noise seed. With a second seed, two steps succeed on 97.2\% of all 500 LIBERO-Goal initial states, against 96.6\% with the first.
Under fresh noise and on LIBERO-Long, two and ten steps differ by at most four points.
Yang~\citep{sbintuitions2026} reports 95.6\% for one step on 500 episodes without naming the checkpoint. Our 93\% with the original checkpoint lies within the sampling error of that figure on 100 states, while 88\% with Optional-IDM does not; a single step is the same at every shift, so the gap lies in the checkpoint or the evaluation.
On LIBERO-Long the $\tau{=}8$ cache behaves as it does on LIBERO-Goal. It keeps every state with the original checkpoint, which caches 44\% of the path at $s{=}5$, and loses 18 points with Optional-IDM, which caches 80\% at $s{=}1$ ($p{=}0.004$ against ten steps).

\begin{table}[!ht]
\centering
\caption{Two against ten uncached steps beyond Table~\ref{tab:frontier}, with one step and the $\tau{=}8$ cache where run. Success is in percent, and the interval is a Tango 90\% interval on the paired difference, in points. The fresh-noise cache column is the default cache at each checkpoint's own shift. The 250-state jitter row ran on the idle machine and repeats every two- and ten-step outcome of the 50-state row on the states the two share.}
\label{tab:app-regimes}
\footnotesize
\begin{tabular}{@{}llrrrrrr@{}}
\toprule
Condition & Ckpt. & states & 10 steps & 2 steps & 2$-$10, Tango 90\% & 1 step & cache $\tau{=}8$ \\
\midrule
LIBERO-Goal, all states & IDM & 500 & 96.4 & 96.6 & $[-1.4, 1.8]$ & -- & -- \\
Camera jitter 1.5\,cm & IDM & 50 & 66 & 66 & $[-9.2, 9.2]$ & 68 & 24 \\
Camera jitter 1.5\,cm & IDM & 250 & 64.8 & 70.0 & $[1.4, 9.3]$ & -- & -- \\
Camera jitter 1.5\,cm & orig. & 50 & 48 & 44 & $[-14.4, 5.9]$ & -- & -- \\
Fresh noise per call & IDM & 50 & 96 & 96 & $[-8.1, 8.1]$ & -- & 26 \\
Fresh noise per call & orig. & 50 & 96 & 92 & $[-11.4, 1.3]$ & -- & 94 \\
LIBERO-Long & IDM & 50 & 98 & 96 & $[-9.7, 5.1]$ & -- & 80 \\
LIBERO-Long & orig. & 50 & 96 & 98 & $[-5.1, 9.7]$ & -- & 100 \\
\bottomrule
\end{tabular}

\end{table}

\subsection{Latency}
\label{app:latency}
Table~\ref{tab:app-latency} gives the mean latency of a call on each machine that timed it, with the fixed cost $F$ of a call in units of the per-step cost $a$. Means average each episode's mean call time, and p95 pools all calls.
The slow-CPU follow-up machine and the shared machine ran the same calls about twice as slowly as the Table~\ref{tab:frontier} machine, the idle machine 1.5 times as slowly, and on the follow-up machine a denoising step costs more relative to the rest of the call.
There, where ten steps cost 2.5 times as much as two, our $\tau{=}8$ cache ran 2.0$\times$ faster than ten steps, close to the 2.13$\times$ that \cthree{} reports on an A100.
Fewer steps also cut the tail: refreshes put the $\tau{=}8$ cache's p95 at 214\,ms (mean 120\,ms), against 95\,ms for two uncached steps. Caching each block separately instead of the whole stack kept attention running, so a reuse call took 146 against 101\,ms. Under upstream CUDA graphs, our cached residuals are overwritten between calls; copying them out of the graph's memory pool on every call should avoid this, at a cost we have not measured.
Table~\ref{tab:app-composed} gives the success and timing of the composed cache ($K{=}2$, step 0 cached, $\tau{=}4$).
Within a run, reuse and refresh calls alternate under the same load, and reuse calls cost 83--87\% of a refresh call on all four machines, a mean saving of 10--13\% at $\tau{=}4$ relative to refresh calls, which also write the cache, rather than to uncached ones.
Refresh calls also store the residuals, and a baseline timed in another process adds that process's own variation: against two uncached steps timed separately, the saving averaged 10\% on the idle machine (5--14\% in single rounds, while its two-step calls varied by 5\% between rounds), 7--9\% on the shared machine and 2\% in two of three interleaved rounds on the follow-up machine.

\begin{table}[!ht]
\centering
\caption{Mean latency per call (ms) and fixed cost $F$ in units of the per-step cost $a$. On the Table~\ref{tab:frontier} machine $F/a$ comes from a least-squares fit over 1--10 steps. On the follow-up machine it comes from the 2- and 10-step calls, because its 1-step arm changed load regime halfway through. The idle machine's row averages its three interleaved rounds.}
\label{tab:app-latency}
\footnotesize
\begin{tabular}{@{}lrrrrrr@{}}
\toprule
Host and checkpoint & 1 step & 2 steps & 4 steps & 10 steps & cache $\tau{=}8$ & $F/a$ \\
\midrule
Table~1 host, IDM & 81 & 94 & 119 & 202 & 116 & 4.9 \\
Table~1 host, orig. & 81 & 92 & 118 & 202 & 120 & 4.8 \\
Table~1 host, orig., CUDA graphs & 79 & 86 & 100 & 143 & -- & 10.0 \\
Follow-up host, IDM (round 1) & 186 & 192 & -- & 477 & 240 & 3.4 \\
Shared host, IDM & 173 & 185 & -- & -- & -- & -- \\
Shared host, orig. & 158 & 190 & -- & -- & -- & -- \\
Idle host, IDM (3 rounds) & -- & 140 & -- & -- & -- & -- \\
\bottomrule
\end{tabular}

\end{table}

\begin{table}[!ht]
\centering
\caption{The composed cache ($K{=}2$, step 0 cached, $\tau{=}4$). \emph{Success}: on the 100 LIBERO-Goal states, where two uncached steps succeed on 95\% (IDM) and 96\% (orig.). The follow-up and idle rounds timed three states per task. \emph{Reuse} and \emph{refresh}: mean call time within the same run. \emph{Saving}: mean saving per call at $\tau{=}4$ implied by the two. \emph{vs.\ 2 steps}: composed-cache latency over that of two uncached steps timed separately on the same machine. The quiet machine had no such baseline.}
\label{tab:app-composed}
\footnotesize
\setlength{\tabcolsep}{4pt}
\resizebox{\linewidth}{!}{\begin{tabular}{@{}lrrrrrr@{}}
\toprule
Host & success (\%) & reuse (ms) & refresh (ms) & reuse/refresh & saving (\%) & vs.\ 2 steps \\
\midrule
quiet A6000, IDM & 94 & 83 & 95 & 0.870 & 10 & -- \\
quiet A6000, orig. & 95 & 83 & 95 & 0.869 & 10 & -- \\
shared A6000, IDM & 94 & 165 & 197 & 0.838 & 12 & 0.93 \\
shared A6000, orig. & 95 & 163 & 196 & 0.835 & 12 & 0.91 \\
follow-up A6000, IDM, round 1 & -- & 178 & 211 & 0.843 & 12 & 0.98 \\
follow-up A6000, IDM, round 2 & -- & 154 & 184 & 0.835 & 12 & 0.80 \\
follow-up A6000, IDM, round 3 & -- & 175 & 210 & 0.835 & 12 & 0.98 \\
idle A6000, IDM, round 1 & -- & 119 & 143 & 0.837 & 12 & 0.88 \\
idle A6000, IDM, round 2 & -- & 122 & 146 & 0.831 & 13 & 0.95 \\
idle A6000, IDM, round 3 & -- & 116 & 139 & 0.832 & 13 & 0.86 \\
\bottomrule
\end{tabular}
}
\end{table}

\subsection{Cache screen}
\label{app:screen}
\begin{figure}[!ht]
\centering
\includegraphics[width=0.6\linewidth]{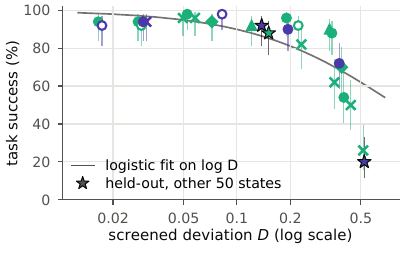}
\caption{Closed-loop success against the screen's deviation $D$ for the 24 $\tau{=}8$ configurations of Table~\ref{tab:app-screen} (markers as in Fig.~\ref{fig:shift}; diamonds: $s{=}2$). Grey: the logistic fit on $\log D$ that made the held-out predictions. Stars: held-out configurations on the other 50 states. Bars: 95\% Wilson intervals.}
\label{fig:app-screen}
\end{figure}

The screen arm of \texttt{probes/fw\_libero.py} runs the uncached policy and, after each call, replays that call's Euler loop from the same initial latent under each configuration (three to nine of ten steps cached at $\tau{=}8$, and eight at $\tau{=}2$, 4 and 0), with the whole-stack residuals the configuration would hold, those recorded on its last refresh call of the same rollout (\texttt{probes/screen\_engine.py}). A replay thus performs the cache's computation for the observation, initial latent and stored residuals of the uncached rollout, not for the states the cache's own actions would reach. On one episode, the replay at the first reuse call matched the action of a closed-loop $\tau{=}8$ cache to the last bit, as the first call matched the cache's refresh; once the cache's actions had moved the robot elsewhere, the two differed by 0.013 at the next call and by up to 0.95 later. The screened rollouts execute only uncached actions and repeated all 400 outcomes of the uncached runs they duplicate. Replaying ten configurations took 2.4 times as long as the uncached call. The predictions P4--P8 are stated in \texttt{repro/experiments/screen.sh}, committed with the analysis before any screening run, and all five held as stated, P4, P7 and P8 by small margins. At $s{=}1$ with eight cached steps, $D$ was 0.401 for Optional-IDM and 0.378 for the original checkpoint (P4). Fresh noise raised $D$ at 4, 6, 7 and 8 cached steps (P5), and at eight cached steps $D$ fell from 0.401 at $s{=}1$ to 0.287, 0.121 and 0.029 at $s{=}2$, 3 and 5 (P6). Over the 24 $\tau{=}8$ configurations of Table~\ref{tab:app-screen}, Spearman's $\rho$ between $D$ and success was $-0.762$, against $-0.738$ for the share (P7), and the bootstrap 90\% interval of the difference over states, $[-0.09, 0.04]$, includes zero. With the refresh intervals of Table~\ref{tab:frontier}, which the share cannot tell apart, $\rho$ was $-0.64$ against $-0.60$ over 31 configurations. A logistic regression on $\log D$ fitted to the 24 predicted the held-out configurations' success with a mean absolute error of 16.4 points, against 16.7 for the share (P8). On the held-out states, the original checkpoint at $s{=}1$ under fresh noise succeeded on 92 and 20\% with six and eight cached steps (96\% uncached), and Optional-IDM at $s{=}2$ on 88\% with seven (98\% uncached). Descriptively, with thresholds chosen after the runs and not among the predictions, all 16 configurations with $D<0.13$ kept at least 90\% success and all six above 0.37 at most 72\%. Under fixed noise at $s{=}1$ with eight cached steps, 27 and 28\% of the two checkpoints' reuse calls inverted the gripper command on all ten executed actions, against at most 6\% under fresh noise (Fig.~\ref{fig:app-screen}).

\begin{table}[!ht]
\centering
\caption{The cache screen (Sec.~\ref{sec:shift}) on the 50 states of Fig.~\ref{fig:shift}. \emph{Cached}: leading steps of ten served from the cache, refreshed every $\tau$ calls ($\tau{=}0$: only when an episode starts). $D$: mean RMS difference between the cached and the uncached action chunk over the ten executed actions, in normalized units. \emph{Success}: the configuration in closed loop on the same states. Below the second rule, configurations never run in closed loop before, on the other 50 states (initial states 5--9 of every task), with the success predicted from $D$ by the logistic fit to the rows above the first rule.}
\label{tab:app-screen}
\footnotesize
\begin{tabular}{@{}lrlrrrrrr@{}}
\toprule
Ckpt. & $s$ & noise & cached & $\tau$ & share (\%) & $D$ & success (\%) & pred.\ (\%) \\
\midrule
IDM & 1 & fixed & 3 & 8 & 30 & 0.017 & 94 & -- \\
IDM & 1 & fixed & 4 & 8 & 40 & 0.028 & 94 & -- \\
IDM & 1 & fixed & 5 & 8 & 50 & 0.052 & 98 & -- \\
IDM & 1 & fixed & 6 & 8 & 60 & 0.190 & 96 & -- \\
IDM & 1 & fixed & 7 & 8 & 70 & 0.344 & 88 & -- \\
IDM & 1 & fixed & 8 & 8 & 80 & 0.401 & 54 & -- \\
IDM & 2 & fixed & 7 & 8 & 54 & 0.072 & 94 & -- \\
IDM & 2 & fixed & 9 & 8 & 82 & 0.390 & 70 & -- \\
IDM & 3 & fixed & 8 & 8 & 57 & 0.121 & 92 & -- \\
IDM & 3 & fixed & 9 & 8 & 75 & 0.333 & 90 & -- \\
IDM & 5 & fixed & 8 & 8 & 44 & 0.029 & 92 & -- \\
IDM & 5 & fixed & 9 & 8 & 64 & 0.222 & 92 & -- \\
IDM & 1 & fresh & 4 & 8 & 40 & 0.050 & 96 & -- \\
IDM & 1 & fresh & 6 & 8 & 60 & 0.231 & 82 & -- \\
IDM & 1 & fresh & 7 & 8 & 70 & 0.355 & 62 & -- \\
IDM & 1 & fresh & 8 & 8 & 80 & 0.516 & 26 & -- \\
IDM & 3 & fresh & 9 & 8 & 75 & 0.439 & 50 & -- \\
IDM & 5 & fresh & 8 & 8 & 44 & 0.058 & 96 & -- \\
orig. & 1 & fixed & 5 & 8 & 50 & 0.030 & 94 & -- \\
orig. & 1 & fixed & 7 & 8 & 70 & 0.194 & 90 & -- \\
orig. & 1 & fixed & 8 & 8 & 80 & 0.378 & 72 & -- \\
orig. & 5 & fixed & 8 & 8 & 44 & 0.017 & 92 & -- \\
orig. & 5 & fixed & 9 & 8 & 64 & 0.083 & 98 & -- \\
orig. & 5 & fresh & 8 & 8 & 44 & 0.031 & 94 & -- \\
\midrule
IDM & 1 & fixed & 8 & 0 & 80 & 0.439 & 8 & -- \\
IDM & 1 & fixed & 8 & 2 & 80 & 0.192 & 96 & -- \\
IDM & 1 & fixed & 8 & 4 & 80 & 0.188 & 82 & -- \\
orig. & 5 & fixed & 8 & 0 & 44 & 0.020 & 92 & -- \\
orig. & 5 & fixed & 8 & 2 & 44 & 0.021 & 90 & -- \\
orig. & 5 & fixed & 8 & 4 & 44 & 0.017 & 92 & -- \\
orig. & 5 & fresh & 8 & 0 & 44 & 0.035 & 90 & -- \\
\midrule
orig. & 1 & fresh & 6 & 8 & 60 & 0.138 & 92 & 86 \\
orig. & 1 & fresh & 8 & 8 & 80 & 0.523 & 20 & 61 \\
IDM & 2 & fresh & 7 & 8 & 54 & 0.151 & 88 & 85 \\
\bottomrule
\end{tabular}

\end{table}

\begin{table}[!ht]
\centering
\caption{Per-episode latent and delayed execution (50 LIBERO-Goal states; Optional-IDM at $s{=}1$, the original checkpoint at $s{=}5$). A and B are success rates, the difference A$-$B is paired by state and given in points with its Tango 90\% interval, and $p$ is an exact McNemar test, uncorrected. \emph{Episode latent}: each episode's first fresh-noise latent kept for all of its calls. \emph{Delay} $d$: the first $d$ of the ten actions executed after a call come from the previous chunk; 5/2 is 5 after refresh calls and 2 after reuse calls. The last block uses the delays of CUDA-graph execution, 2 for two steps and 3 for ten.}
\label{tab:app-latent-delay}
\footnotesize
\setlength{\tabcolsep}{4pt}
\begin{tabular}{@{}llrrrrr@{}}
\toprule
A & B & A (\%) & B (\%) & A$-$B & Tango 90\% & $p$ \\
\midrule
IDM, 6 of 10 cached, episode latent & fresh noise & 96 & 82 & +14 & $[3.5, 25.4]$ & 0.065 \\
IDM, 6 of 10 cached, episode latent & fixed latent & 96 & 96 & 0 & $[-8.1, 8.1]$ & 1.00 \\
IDM, 8 of 10 cached, episode latent & fresh noise & 64 & 26 & +38 & $[23.3, 51.2]$ & 0.0002 \\
IDM, 8 of 10 cached, episode latent & fixed latent & 64 & 54 & +10 & $[0.1, 20.8]$ & 0.18 \\
\midrule
IDM, 2 steps, delay 2 & 10 steps, delay 5 & 96 & 86 & +10 & $[1.6, 20.0]$ & 0.12 \\
IDM, 2 steps, delay 2 & 2 steps, no delay & 96 & 94 & +2 & $[-5.1, 9.7]$ & 1.00 \\
IDM, 10 steps, delay 5 & 10 steps, no delay & 86 & 96 & -10 & $[-20.8, -0.1]$ & 0.18 \\
IDM, 2 steps, delay 5 & 2 steps, delay 2 & 82 & 96 & -14 & $[-24.7, -4.9]$ & 0.039 \\
IDM, 10 steps, delay 5 & 10 steps, delay 2 & 86 & 96 & -10 & $[-19.2, -4.4]$ & 0.062 \\
IDM, 10 steps, delay 2 & 2 steps, delay 2 & 96 & 96 & 0 & $[-6.8, 6.8]$ & 1.00 \\
\midrule
orig., cache $\tau{=}8$, delay 5/2 & 2 steps, delay 2 & 88 & 94 & -6 & $[-14.1, -0.6]$ & 0.25 \\
orig., cache $\tau{=}8$, delay 5/2 & same, no delay & 88 & 92 & -4 & $[-13.4, 4.8]$ & 0.69 \\
orig., 2 steps, delay 2 & 2 steps, no delay & 94 & 94 & 0 & $[-6.8, 6.8]$ & 1.00 \\
\midrule
IDM, 2 steps, delay 2 & 10 steps, delay 3 & 96 & 90 & +6 & $[0.6, 14.1]$ & 0.25 \\
IDM, 10 steps, delay 3 & 10 steps, no delay & 90 & 96 & -6 & $[-16.0, 3.2]$ & 0.45 \\
IDM, 10 steps, delay 4 & 10 steps, delay 2 & 94 & 96 & -2 & $[-9.7, 5.1]$ & 1.00 \\
orig., 2 steps, delay 2 & 10 steps, delay 3 & 94 & 94 & 0 & $[-6.8, 6.8]$ & 1.00 \\
orig., 10 steps, delay 3 & 10 steps, no delay & 94 & 94 & 0 & $[-8.1, 8.1]$ & 1.00 \\
\bottomrule
\end{tabular}

\end{table}

\begin{table}[!ht]
\centering
\caption{Successes of 10 per LIBERO-Goal task behind Table~\ref{tab:frontier} (one RTX A6000): two and ten uncached steps and the $\tau{=}8$ cache, Optional-IDM (IDM) at $s{=}1$ and the original checkpoint at $s{=}5$.}
\label{tab:app-per-task}
\footnotesize
\begin{tabular}{@{}rlrrrrrr@{}}
\toprule
& & \multicolumn{3}{c}{IDM} & \multicolumn{3}{c}{orig.} \\ \cmidrule(lr){3-5} \cmidrule(l){6-8}
& Task & 2 steps & 10 steps & $\tau{=}8$ & 2 steps & 10 steps & $\tau{=}8$ \\
\midrule
0 & open the middle drawer of the cabinet & 10 & 10 & 9 & 10 & 10 & 10 \\
1 & put the bowl on the stove & 9 & 10 & 8 & 10 & 10 & 10 \\
2 & put the wine bottle on top of the cabinet & 9 & 7 & 10 & 10 & 9 & 8 \\
3 & open the top drawer and put the bowl inside & 8 & 9 & 3 & 9 & 9 & 9 \\
4 & put the bowl on top of the cabinet & 10 & 10 & 9 & 10 & 10 & 10 \\
5 & push the plate to the front of the stove & 10 & 10 & 1 & 9 & 8 & 8 \\
6 & put the cream cheese in the bowl & 10 & 10 & 4 & 10 & 10 & 10 \\
7 & turn on the stove & 10 & 10 & 10 & 10 & 10 & 10 \\
8 & put the bowl on the plate & 10 & 10 & 4 & 10 & 10 & 10 \\
9 & put the wine bottle on the rack & 9 & 9 & 5 & 8 & 9 & 9 \\
\midrule
& total & 95 & 95 & 63 & 96 & 95 & 94 \\
\bottomrule
\end{tabular}

\end{table}

\subsection{Identity gate}
\label{app:gate}
The experiment driver runs the gate of Sec.~\ref{sec:setup} once per collection, before its first arm, and stops the collection if the replayed actions differ from uncached ones by 0.5\% or more in relative L2 error. The gates shipped in \texttt{results/budget\_shift} record errors of at most 0.17\%, though not every directory there has its own.
The eight residuals of the default cache take 0.5\,MB in bf16, and each refresh also keeps those of the two uncached steps, which no call reuses.

\subsection{Reproducibility}
\label{app:repro}
The planned re-runs, 47 comparisons over seven A6000 machines and five revisions of the harness, changed none of 2,730 paired outcomes and 14 lengths. Over all 249 pairs of runs that share a configuration but not a machine (thirteen machines, eleven revisions), 7 of 11,725 outcomes changed, on five initial states, and 117 lengths.
Latency did not transfer: the same uncached calls ran 21--41\% slower on one machine and 1.95--2.36 times slower on two others than on the Table~\ref{tab:frontier} machine.
\end{document}